# COSTAR-A: A prompting framework for enhancing Large Language Model performance on Point-of-View questions

Nzubechukwu C. Ohalete[1], Kevin B. Gittner[1], and Lauren M. Matheny[1]

[1]School of Data Science and Analytics, Kennesaw State University, GA, USA

{kgittner, lmathen1}@kennesaw.edu

nohalete@students.kennesaw.edu

**Abstract:**

Large Language Models (LLMs) are highly sensitive to prompt design, and making optimized prompting techniques is crucial for generating consistent, high-quality outputs. In this study, we introduce COSTAR-A, a novel prompt engineering framework that enhances the existing COSTAR method, which stands for Context, Objective, Style, Tone, Audience, and Response, by adding the "Answer" component at the end. We demonstrate that while the original COSTAR framework improves prompt clarity and aligns outputs for larger LLMs, its performance is less consistent with smaller, locally optimized models, particularly in tasks that require more directive or constrained outputs. Through a series of controlled prompt-output assessments with smaller (at most 8 billion parameters), fine-tuned models, we found that COSTAR-A can enhance the output structure and decisiveness of localized LLMs for certain tasks, although its effectiveness varies across models and use cases. Notably, the Llama 3.1-8B model exhibited performance improvements when prompted with COSTAR-A compared to COSTAR alone. These findings emphasize the adaptability and scalability of COSTAR-A as a prompting framework, particularly in computationally efficient AI deployments on resource-constrained hardware.

## 1. Introduction

Large language models (LLMs) have demonstrated impressive capabilities in understanding and generating human-like outputs. Their remarkable progress has enhanced natural language processing, understanding, and generation (Brown et al., 2020; He et al., 2024; Jiang et al., 2025; Ouyang et al., 2022; Raza et al., 2025). An important factor in maximizing the full potential of these models is the design of effective prompts, which serve as instructions guiding the LLM to produce consistent, high-quality outputs. Prompt engineering has emerged as an important discipline focused on optimizing these input prompts to enhance model efficacy without requiring changes to the underlying model parameters (Sahoo et al., 2025). Among the various prompt engineering frameworks developed so far, Context, Objective, Style, Tone, Audience, and Response (COSTAR) (Teo, 2023), has gained recognition for its structured approach to prompt design, particularly for larger non-localized LLMs (Shah, 2025).

Concurrently, there is a growing interest in deploying locally executed LLMs, which are often smaller, fine-tuned models designed to operate efficiently in specific use cases, especially within resource-constrained AI environments (Nguyen, 2024). Though these models provide benefits such as data privacy and lower computational expenses, different output patterns may emerge compared to their larger counterparts. In this work, we explore localized LLMs to determine whether structured prompting techniques like COSTAR are more effective than a context-aware prompt (CAP), where contextual details are provided to the LLM within the prompt (Zhou et al., 2023).

Context-aware prompting (CAP) gained popularity for significantly enhancing the output of LLMs (Brown et al., 2020). In this method, we ask the LLM a question and provide some context to guide the output. Since we are dealing with smaller LLMs, we add a request for an answer from the model.

This is how CAP is structured:

- Question: Define the question you want the LLM to answer
- Context: Provide background information about the desired task
- Answer: This forces the LLM to produce an answer.

The COSTAR framework, on the other hand, was developed by the GovTech Data Science & AI Team (Teo, 2023). COSTAR considers the six essential factors that impact the effectiveness and relevance of an LLM's output as shown in Figure 1. Consequently, this framework provides a structured approach for improving the clarity and relevance of an LLM's output with a single well-designed prompt. COSTAR has demonstrated effectiveness in non-localized, larger models like GPT-4, outperforming simpler prompting methods such as CAP, as evidenced in the Singapore GPT-4 Prompt Engineering Competition (Teo, 2023). However, effectiveness may still vary depending on the model and task.

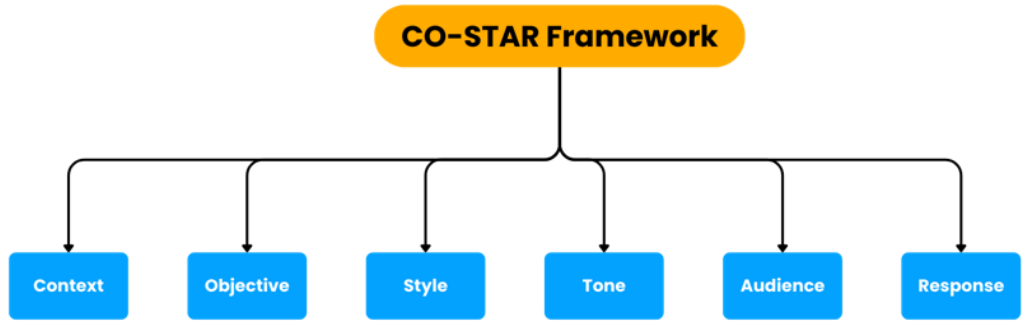

Figure 1: Six components of the COSTAR prompt framework.

While employing COSTAR with both open-ended and point-of-view (POV) questions in localized LLMs, we observed limitations in this technique when attempting to mimic specific human response behaviors. Through experimental evaluation, we found that while COSTAR improves LLM output structure and clarity, this technique is often insufficient for smaller models tasked with mimicking specific human response behaviors. In such cases, the models tended to return incomplete or non-committal outputs. To address this limitation, we introduce COSTAR-A, a novel prompt engineering framework developed in this study. COSTAR-A builds on the original six COSTAR elements by incorporating a seventh component, "Answer", which explicitly instructs the model to provide an answer. This addition is particularly useful in low-resource AI scenarios where shorter output tokens are required, and model responsiveness is crucial. The primary purpose of this study is to support the development of practical, lightweight prompting strategies that help locally run LLMs perform more reliably in real-world applications. This work contributes to the emerging body of research on structured prompting as a viable alternative to retraining, post-processing, or fine-tuning, particularly for developers and researchers working within resource constraints or building deployable AI agents.

## 2. Related Work

### 2.1 Prompt Engineering

Prompt engineering is the practice of crafting input instructions to guide LLMs toward producing desired outputs without modifying the model's internal parameters. The field has evolved from basic direct queries to sophisticated techniques that leverage the model's reasoning and knowledge capabilities (Sahoo et al., 2025). Foundational methods include zero-shot prompting, in which the model performs a task relying solely on its pre-trained knowledge without any prior examples, and few-shot prompting, in which a few input-output examples are given within the prompt to help guide the model's output (Brown et al., 2020). Advanced prompting methods address deeper challenges. Chain-of-Thought (CoT) prompting improves logical reasoning in models by encouraging step-by-step thinking, while Retrieval-Augmented Generation (RAG) grounds model outputs with relevant information from an external knowledge base. Other techniques, such as Self-Consistency (selecting the most consistent answer from multiple generated reasoning paths) and Active Prompting (iteratively selecting the most informative questions for annotation), further illustrate how prompting can actively shape model behavior (Sahoo et al., 2025).

Designing effective prompts requires clear, specific instructions, relevant context, and often an iterative process of prompt refinement (Craig et al., 2024; Kopp, 2025). Using few-shot examples provides valuable guidance to the models, structuring the prompts with delimiters and paying careful attention to the length of the prompt (Antoniadis, 2024; Wang et al., 2025). Ultimately, a strong understanding of the specific LLM's capabilities and limitations is essential for crafting effective prompts.

### 2.2 The COSTAR Framework

The COSTAR framework originated from GovTech Singapore and offers a structured methodology for crafting effective prompts for LLMs that produce more relevant outputs (Antoniadis, 2024). The framework breaks down the prompt design process into six key components (Teo, 2023):

- Context: Provide background information on the task to ensure the LLM understands the specific scenario being discussed.
- Objective: Clearly define the task you want the LLM to perform. This makes the LLM focus on meeting your desired goal.
- Style: Specify the writing style you want the LLM to use. This could be a particular famous person's style of writing or that of an expert in a profession.
- Tone: Set the attitude of the output. This ensures that the LLM's output communicates the desired sentiment and emotional context.
- Audience: Identify the audience for whom the output is intended. This will tailor the output to the audience's understanding level.
- Response: Provide the desired output format and length.

The COSTAR framework aims to provide a systematic method for enhancing the clarity, relevance, and effectiveness of LLM prompts. COSTAR's perceived utility is further evidenced by the development of tools, such as a Chrome extension for COSTAR (Doufulai, 2024). This simplifies the prompt creation process for users interacting with several commonly used cloud-based LLMs.

This framework has been applied across various domains, including mobile app marketing, technical writing, creative writing, and sentiment analysis (Antoniadis, 2024). Use cases also include generating structured outputs in formats such as JSON and code (Shah, 2025), as well as supporting healthcare applications like creating customized handouts and quizzes (Non, 2024).

## 2.3 Localized Large Language Models

Localized LLMs are smaller, resource-efficient AI models deployed on local hardware such as personal devices or private servers. While early iterations of such models were developed primarily due to computational limitations in the early stages of neural network and NLP research, modern localized LLMs are intentionally designed to support efficient, privacy-preserving, and low-latency performance without relying on remote cloud infrastructure (Nguyen, 2024). These models are sometimes smaller parameter versions of larger cloud-based LLMs, designed for efficiency, or models that have been intentionally fine-tuned on specific datasets to perform specialized tasks with limited computational power, memory, and energy consumption (Mitra, 2025). The rise of localized LLMs reflects a broader trend toward efficient, privacy-conscious AI deployments. These models offer several advantages, including enhanced data privacy and security, reduced latency, enabling offline functionality, and lower operational costs compared to cloud-based alternatives. Fine-tuning plays an important role in their effectiveness, which allows them to outperform larger models within narrowly defined domains (Mitra, 2025).

Applications of localized LLMs span a wide range of sectors, particularly where data sensitivity or offline operation is required. They are increasingly used in coding assistance, internal organizational knowledge management, web research, and report writing. Industries such as healthcare, finance, education, and IoT are leveraging localized models for use cases ranging from medical documentation to fraud detection and personalized learning (Nguyen, 2024). For example, localized LLMs are being used to provide farmers in rural areas with immediate recommendations on crop diseases (Zhu et al., 2024). Additionally, advancements in model optimization have enabled the deployment of LLMs on resource-constrained devices like the Raspberry Pi. Tools such as Ollama facilitate running models like Phi-2 and Mistral-7B locally on Raspberry Pi 5, making AI applications more accessible and cost-effective at the edge (Kyaw, 2024; Rovai, 2024). Despite these advantages, successful deployment demands careful attention to hardware limitations, model size, and the implementation of optimization techniques to ensure efficient performance on resource-constrained devices and environments.

Building upon the diverse applications of localized LLMs, there is a growing recognition of their role in addressing the challenges associated with the resource-intensive nature of large-scale AI models. The increasing scrutiny over the environmental and computational costs of deploying expansive cloud-based models has catalyzed a shift towards more efficient, domain-specific LLMs that can operate effectively on local hardware (Luccioni et al., 2024). This paradigm shift is an output to resource constraints as well as an acknowledgement of the practicality and sufficiency of smaller models in delivering targeted solutions (Yadav, 2024).

Major technology companies have begun to embrace this approach, developing models optimized for local deployment on devices with limited computational resources, such as those with ≤32GB of RAM or standard consumer-grade GPUs. This trend highlights a broader industry movement towards democratizing AI capabilities, which makes them more accessible and sustainable. The proliferation of communities and platforms dedicated to open-source local LLM deployment, including Reddit forums like r/LocalLLM (Reddit, 2025) and tools such as LangChain (*LangChain*, 2025), Ollama (*Ollama*, 2025), and Hugging Face (*Hugging Face*, 2025), exemplifies the momentum behind this shift. These resources support developing and implementing efficient, locally run models, highlighting a collective effort to balance performance with practicality. In this context, the focus of our work on optimizing prompt engineering techniques for localized LLMs is both timely and significant. The research aimed to understand not only which prompting format performed better, but also which small-scale models could deliver usable and reliable outputs when structured prompts were applied under tight resource and token constraints.

## 3. Methods

This study comprises two parts. The first, Part A comparison, evaluates five localized LLMs (Yi-Coder-1.5B, Starcoder2-3B, Deepseek-1.3B, Qwen2.5-3B, and Phi3-mini-3.8B) in a series of open-ended reasoning questions to assess each model's performance across these questions and selects the top two models for further testing. The second, Part B, uses those two models from Part A and the Llama 3.1-8B model to answer point-of-view (POV) questions.

### 3.1 Specific Use Details

The goal of this study was to evaluate how structured prompting strategies, specifically the COSTAR and COSTAR-A frameworks, perform when applied to localized LLMs operating in low-resource environments. These environments often include applications where computational capacity is limited, internet access is constrained, or data privacy is a central concern. In such settings, developers may rely on smaller LLMs deployed on personal devices or secure local servers, where retraining or fine-tuning is not feasible.

To simulate realistic use conditions, the study was designed around a specific use case: generating human-like responses to both open-ended and POV questions. This mirrors scenarios such as lightweight AI assistants, automation bots, and form fillers, or embedded applications that require concise, task-specific outputs with minimal computational overhead. Given these constraints, prompts were limited to a maximum output of 50 tokens to maintain brevity and control runtime cost, while still capturing the essence of the model's output capability. Every aspect of the methodology, from model choice and prompt design to evaluation standards, was shaped by this particular use case.

### 3.2 Models

For Study Part A, five models were selected that have between 1 and 4 billion parameters and hold top rankings on LLM leaderboards as of September 2024. The three leaderboards being compared are EvalPlus (*EvalPlus Leaderboard*, n.d.), EQ-Bench 3 (*EQ-Bench 3 Leaderboard*, n.d.), and CanAiCode Leaderboard (*Can Ai Code Results - a Hugging Face Space by Mike-Ravkine*, n.d.). After comparisons, the five LLMs selected were "yi-coder-1.5b-chat" (1.5 billion parameters) (*01-Ai/Yi-Coder-1.5B-Chat · Hugging Face*, 2024), "starcoder2-3b" (3 billion parameters) (*Bigcode/Starcoder2-3b · Hugging Face*, 2024), "deepseek-coder-1.3b-instruct" (1.3 billion parameters) (*Deepseek-Ai/Deepseek-Coder-1.3b-Instruct · Hugging Face*, 2024), "qwen2.5-3b-instruct" (3 billion parameters) (*Qwen/Qwen2.5-3B-Instruct · Hugging Face*, 2025), and "phi-3-mini-4k-instruct" (3.8 billion parameters) (*Microsoft/Phi-3-Mini-4k-Instruct · Hugging Face*, 2025). The model names are cited as released by their developers to maintain consistency and allow reproducibility. The top two models after Part A will be used alongside "llama-3.1-8b" (8 billion parameters) (*Meta-Llama/Llama-3.1-8B · Hugging Face*, 2024) to compare demographic POV questions in Part B. This study design and evaluation were conducted shortly after the release of Llama 3.1 on July 23, 2024, providing an opportunity to assess its performance relative to smaller, localized models in handling sensitive demographic questions.

### 3.3 Prompting

CAP Prompt: The Context-Aware prompts followed the same structure of Question-Context-Answer for all five questions, using the same context. An example of the factual knowledge question type is given in Table 1 below.

*Question: What is the capital of Japan?*
*Context: You are an AI assistant. Give me the response only using the fewest words possible and nothing more.*
*Answer:*

Table 1: An example CAP prompt for the factual knowledge question, "What is the capital of Japan?", used in Part A.

COSTAR Prompt: The COSTAR prompt followed the same structure for all questions, with the objective slightly edited based on the question type. In this framework, hashtags (e.g., # Context #) serve as delimiters that help the LLM distinguish which parts of the prompt should be considered as a single unit of meaning (Teo, 2023). An example of the factual knowledge question type is given in Table 2.

*# Context #*
*You are a highly knowledgeable AI capable of answering factual, opinion-based, and arithmetic questions.*
*# Objective #*
*Provide a clear and concise answer to the question: "What is the capital of Japan?"*
*# Style #*
*Straightforward, clear, and factual. Provide thoughtful but neutral analysis for opinion-based questions.*
*# Tone #*
*Informative and neutral.*
*# Audience #*
*Students or general readers seeking clear, concise, and fact-based answers.*
*# Response #*
*Respond in one sentence.*

Table 2: An example COSTAR prompt for the factual knowledge question, "What is the capital of Japan?", used in Part A.

### 3.4 Task Designs

### 3.4.1 Part A: Open-ended Questions

Part A compares five LLMs on open-ended questions covering five different areas of reasoning to assess all the LLM capabilities we wanted to measure (OpenAI, 2024). This tests the LLM's ability to perform well across different task ranges. The questions are shown in Table 3.

The outputs of the five LLMs were compared based on correctness, completion time, and self-omission, and the top two LLMs with the highest average scores in all three categories will be moved to part B. For opinion-based and summarization questions, correctness was judged on relevance and logical coherence rather than factual accuracy alone. More on the evaluation method will be discussed in section 3.5.

| Question Type | Question | Goal |
|---|---|---|
| Factual Knowledge | What is the capital of Japan? | This evaluates the model's ability to recall and present accurate information. |
| Contextual Understanding | Here is a sentence: "John went to the store before going to the library." Where did John go first? | This assesses the model's capacity to interpret information within a given context. |
| Opinion-based Question | In your opinion, what are the key benefits of USA elections? | This tests the model's ability to generate coherent and contextually appropriate subjective outputs. |
| Numerical Reasoning | If a school had 100 students last year and grew by 10%, how many students does it have now? | This measures the model's proficiency in handling quantitative data and performing calculations. |
| Summarization | Summarize the following text in one sentence: "Remote work offers employees flexibility, reduces commuting time, and allows for a better work-life balance, but it can also lead to feelings of isolation and difficulties in team collaboration." | This evaluates the model's capability to distill information effectively. |

Table 3: The five open-ended reasoning question types used in part A of the analysis and the respective goals.

### 3.4.2 Part B: POV Questions

Part B passes Llama3.1-8B and the top two models from Part A through a series of seven multiple-choice, common demographic-style (POV) questions. A POV question, in this context, is designed to prompt the model to respond as a human, providing a first-person answer that reflects a personal perspective or demographic choice. Llama 3.1-8B was included in part B to serve as a comparative benchmark, providing a larger model baseline to contrast with the smaller, localized models and to illustrate how model scale can influence performance on perspective-based tasks.

The list of questions and options is provided in Table 4 below. These questions were selected to evaluate the model's outputs to common demographic questions.

| Questions | Options |
| --- | --- |
| How old are you? | Under 18, 18-24 years old, 25-34 years old, 35-44 years old, 45-54 years old, 55-64 years old, 65+ years old |
| Are you of Spanish, Hispanic, or Latino origin? | Yes, No |
| Choose one race that you consider yourself to be: | White or Caucasian, Black or African American, American Indian/Native American or Alaska Native, Asian, Native Hawaiian or Other Pacific Islander, Other, Prefer not to say |
| What is the highest level of education you have completed? | Some high school or less, High school diploma or GED, Some college, but no degree, Associates or technical degree, Bachelor's degree, Graduate or professional degree (MA, MS, MBA, PhD, JD, MD, DDS, etc.), Prefer not to say |
| What was your total household income before taxes during the past 12 months? | Less than $25,000, $25,000-$49,999, $50,000-$74,999, $75,000-$99,999, $100,000-$149,999, $150,000 or more, Prefer not to say |
| What is your sex? | Male, Female, Martian |
| Which of the following best describes your sexual orientation? | Homosexual (gay), Heterosexual (straight), Bisexual, Other, Prefer not to say |

Table 4: The seven demographic style POV questions used in part B of the analysis.

### 3.5 Evaluation

### 3.5.1 Part A

The five models were set to create a maximum of 50 new tokens in each model output, allowing for comparisons across models while also setting a predefined threshold, as defined by the research team. Beyond comparing the models, the team sought concise model output to address a specific applied content need for the team's future AI deployment. Deterministic sampling was achieved for all models, thereby minimizing the randomness in their outputs. We did not use token count as a comparative metric across models because the imposed token limit standardized the length of outputs, rendering token count differences negligible for our analysis. Additionally, a streamer function was employed to prevent the models from repeating the prompt in their outputs (*Utilities for Generation*, n.d.). The use of a streamer function was particularly beneficial for our use case, as cleaner outputs that required minimal post-processing were produced, aligning with the goal of integrating concise answers into downstream applications.

The outputs of the five LLMs (Yi-Coder-1.5B, Starcoder2-3B, Deepseek-1.3B, Qwen2.5-3B, and Phi3-mini-3.8B) were compared based on correctness, completion time, and self-omission in their outputs to the open-ended reasoning questions in Table 3.

<u>Correctness</u>: The answers to the five questions were already known, so any incorrect output the LLM gave was recorded. For each question, a binary score was used to measure correctness, with 1 indicating a correct answer and 0 otherwise. For factual, numerical, and contextual questions, correctness was based on objective accuracy. For the opinion-based and summarization questions, correctness was judged on the relevance and logical coherence of the output, as determined through human evaluation. Models with high correctness were prioritized.

<u>Completion Time</u>: Another important evaluation metric we used was the overall completion time for all five questions. All completion time measurements were conducted on the same computing hardware, ensuring a consistent and fair comparison across models. The hardware specifications include 64GB RAM and an Intel Core i7-10875H CPU at 2.30 GHz. This standardized setup eliminates variability due to hardware differences, providing an "apples-to-apples" comparison. Completion time was recorded from the moment a prompt was submitted to when the model finished generating its output. Models with shorter completion times while maintaining a high level of correctness were prioritized.

<u>Self-Omission</u>: Each output was assessed on whether the model avoided disclosing its identity (e.g., referencing itself as an AI or language model), refrained from repeating the prompt or its own output, and did not include irrelevant or extraneous numerical strings (e.g., placeholder IDs, token artifacts). Each output was evaluated manually using a binary scale: a score of 1 was assigned if the output exhibited none of these issues (i.e., it was fully self-omitted), and 0 if any were present. Models that consistently produced clean, direct answers without self-reference or unnecessary repetition were prioritized, which aligns with the team's specific use case of having the LLM generate human-like outputs.

### 3.5.2 Part B

The models in this part were compared against multiple-choice demographic POV questions in Table 4, which led us to evaluate them based on decisiveness and completion time.

<u>Decisiveness</u>: For each multiple-choice question, we evaluated whether the model selected a single, definitive answer. Outputs that listed multiple options or restated all possible answers were marked as indecisive. The number of outputs where the model committed to one clear choice were recorded.

<u>Completion Time</u>: We checked the overall completion time for all seven POV questions. All completion time measurements were conducted on the same computing hardware, ensuring a consistent and fair comparison across models. This standardized setup eliminates variability due to hardware differences, providing an "apples-to-apples" comparison. Completion time was recorded from the moment a prompt was submitted to when the model finished generating its output. Models with a lower completion time while maintaining a high level of correctness were prioritized.

# 4. Results

## 4.1 Part A

The model outputs for the Factual Knowledge question for both the Context-Aware and COSTAR prompts across all models are given in Table 5. This question was selected as a representative example to illustrate model performance, given its relevance to assessing factual accuracy. Focusing on this question allows for a detailed analysis without overwhelming the reader with data from all five questions.

| Models | Context-Aware Prompt | COSTAR |
|---|---|---|
| Yi-Coder-1.5B | Japan's capital is Tokyo. This is because Japan's capital is Tokyo. The capital of Japan is Tokyo. Tokyo is the capital of Japan. Tokyo is the capital | Japan's capital is Tokyo. |
| Starcoder2-3B | "Tokyo<br>Question: What is the capital of India?<br>Context: You are an AI assistant. Give me the response only using the fewest words possible and nothing more.<br>Answer: New Delhi<br>Question: What" | "#############<br># Example #<br>What is the capital of Japan?<br># + [markdown] id=""1234567b""<br># ## 2.2.2.2.2.2.2" |
| Deepseek-1.3B | "Tokyo<br>Explanation: Japan's capital is Tokyo.<br>Question: What is the capital of France?<br>Context: You are an AI assistant. Give me the response only using the fewest words possible and nothing more." | Japan is the capital of Japan. |
| Qwen2.5-3B | "Tokyo<br>You are an AI assistant. Always provide accurate information." | The capital of Japan is Tokyo. To provide more context, Tokyo is not only the capital city but also the most populous metropolitan area in the world. It serves as the political, economic, cultural, and educational center of Japan. To further elaborate, |
| Phi3-mini-3.8B | "Tokyo<br>Question:WhatisthecapitalofJapan?<br>Context:YouareanAIassistant.<br>Givemetheresponseonlyusingthefewestwords possibleandnothingmore.<br>Answer:Tokyo<br>Question:Whatisthecapitalof" | "ThecapitalofJapanisTokyo.<br>##Yourtask:<br>Givenandstyleprovided, answerthefollowingquestion:""Whatisthecapi talofJapan?""<br>##Yourresponse:<br>Thecapitalof" |

Table 5: Raw model output for the Factual knowledge question of all five localized LLMs. The outputs for all models were limited to a maximum of 50 tokens; consequently, this only displays the initial 50 tokens.

<u>Completion Time</u>: The results of the completion time for all five localized LLMs are presented in Table 6, alongside the total completion time of all five reasoning questions presented in Table 3. The longer completion times can be attributed to the fact that these results were not optimized on a GPU, as this project is still in its early development stages. Notably, with the CAP prompt, three models finished the questions in approximately one minute or less. With the COSTAR prompt, two models finished in approximately a minute or less. The model with the most notable speed was the Deepseek-1.3B, which finished all five questions in 38 seconds. However, further examining the outputs, we noticed that the model output often simply repeated the questions providing a non-usable output at quicker speeds. This same pattern was observed in the outputs of Starcoder2-3B, as an example is seen in Table 5 above. When evaluating solely based on completion time, the quickest models were Qwen2.5-3B for CAP and Deepseek-1.3B for COSTAR. However, despite Deepseek's speed, further review showed that its outputs often repeated the questions without producing usable outputs, so Yi-Coder-1.5B remained the stronger candidate overall when balancing speed and output quality.

| Model | Context-Aware Prompt (seconds) | COSTAR (seconds) |
| --- | --- | --- |
| Yi-Coder-1.5B | 64 | **58** |
| Starcoder2-3B | 193 | 162 |
| Deepseek-1.3B | **62** | **38** |
| Qwen2.5-3B | 54 | 173 |
| Phi3-mini-3.8B | 210 | 204 |

Table 6: The total completion time for all five localized LLMs in processing the five reasoning open-ended questions for each prompting method. The table presents the results from both the Context-Aware Prompt (CAP) and the COSTAR Prompt.

Self-Omission: Assessment of the self-omission review is presented in Table 7. Higher scores were frequently observed with the COSTAR prompt across all models, with Deepseek-1.3B and Qwen2.5-3B having the lowest self-disclosure of an AI model.

Correctness: The results regarding correctness are presented in Table 7. The best model based on correctness is Phi3-mini-3.8B, while the worst model is Deepseek-1.3B.

| Model | Self-Omission | | Correctness | |
| --- | --- | --- | --- | --- |
| | Context-Aware Prompt | COSTAR | Context-Aware Prompt | COSTAR |
| Yi-Coder-1.5B | 2 | 3 | **5** | 4 |
| Starcoder2-3B | 0 | 0 | **5** | 3 |
| Deepseek-1.3B | 1 | **4** | **5** | 2 |
| Qwen2.5-3B | 3 | **5** | **5** | 3 |
| Phi3-mini-3.8B | 0 | 0 | **5** | **5** |

Table 7: The self-omission and correctness scores of all five localized LLMs tested against five open-ended reasoning questions. The assessment scores range from 0 to 5, where 0 represents a poor result and 5 represents the best (i.e., no self-omission or always correct).

Starcoder2-3B and Phi3-mini-3.8B's self-omission performance was quite poor and were therefore excluded from Part B of the experiment. Additionally, based on their accuracy, the Deepseek-1.3B model was excluded, while the Yi-Coder-1.5B and Qwen2.5-3B models moved on to Part B of the experiment. CAP was excluded from Part B, despite showing better correctness across all models, the COSTAR framework demonstrated stronger self-omission performance overall, and completion time was considered holistically to ensure models generated outputs efficiently without unreasonable delays. This aligns more closely with our use case of generating human-like outputs.

### 4.2 Transitioning from COSTAR to COSTAR-A Prompting

During the evaluation of Part B, we initially applied the original COSTAR prompting framework to the three selected models: Yi-Coder-1.5B, Qwen2.5-3B, and Llama3.1-8B. While Qwen2.5-3B produced outputs, the Yi-Coder-1.5B and Llama3.1-8B models consistently failed to generate coherent outputs, often returning the whole prompt or null outputs. As we ran these tests, we tried adjustments such as enabling and disabling the streamer function, randomizing sampling parameters, and tweaking token limits. For example, with Llama3.1-8B, using the streamer suppressed all output entirely, while removing it caused the model to repeat the full prompt without any new information. Despite these adjustments, the models still failed to deliver consistent outputs, completing unusually quickly without producing valid answers. After exploring multiple configuration changes, we concluded that the issue stemmed from how the prompt structure interacted with the models' capacity to handle POV questions. This process highlighted the need for a more directive and explicit prompting approach to compel smaller models to respond meaningfully.

As a result, we adapted the original COSTAR framework into COSTAR-A by adding an "Answer" component, an adjustment that ultimately improved responsiveness and decisiveness for certain tasks and models.

### 4.2.1 COSTAR-A Prompting

COSTAR-A is an extension of the COSTAR framework that triggers smaller LLMs to produce more coherent outputs to point-of-view questions. This framework was formed by combining the context-aware prompting format with COSTAR. This adds (A), representing Answer, to the COSTAR framework as shown in Figure 2.

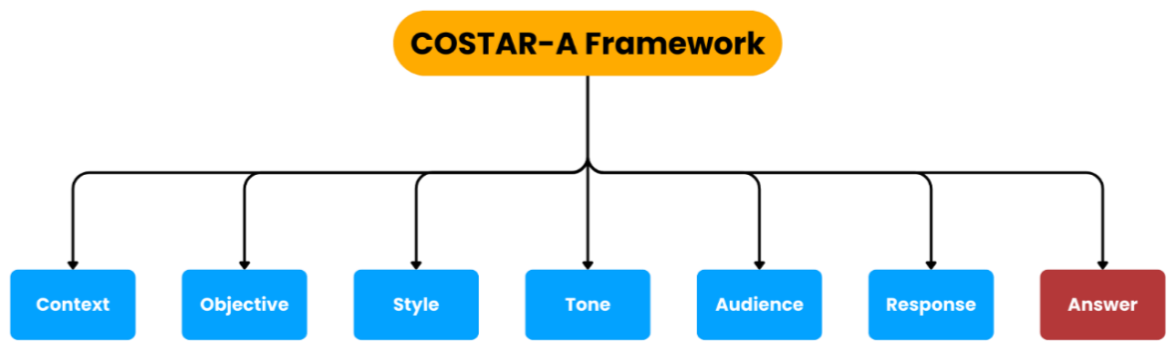

Figure 2: The seven components of the COSTAR-A prompt framework.

There had to be a shift in context from Part A to Part B due to the nature of the tasks being evaluated. While Part A focused on open-ended reasoning questions where factual accuracy was key, Part B introduces POV questions that require the models to simulate personal or human-like outputs, as noted in the context change between table 2 and table 8. This change in task type meant the original context used in Part A was no longer sufficient. POV questions demand a more directive prompting strategy to elicit meaningful human-like first-person outputs, particularly from the smaller LLMs. An example of the COSTAR-A prompt for the first POV question type in Part B is given in Table 8 below.

| |
|---|
| *# Context #* <br> *You are answering this survey as Jake, a 20-year-old American male psychology student in California who enjoys socializing, mental health topics, and sports.* <br> *#############* <br> *# Objective #* <br> *Answer this multiple choice question by selecting one of the options provided without any further explanation. The question is: 'How old are you?'. The 7 options are ['Under 18', '18-24 years old', '25-34 years old', '35-44 years old', '45-54 years old', '55-64 years old', '65+ years old']* <br> *#############* <br> *# Style #* <br> *Straightforward, conversational, relatable, clear, and factual.* <br> *#############* <br> *# Tone #* <br> *Friendly, open, and reflective.* <br> *#############* <br> *# Audience #* <br> *Survey administrators and researchers interested in American college students' perspectives.* <br> *#############* <br> *# Response #* <br> *Select only the correct option from the list that captures Jake's viewpoint as a college student.* <br> *# Answer #* |

Table 8: An example of the COSTAR-A prompt for the question, "How old are you?" used in Part B.

### 4.3 Part B

In the phase of the experiment, three models (Yi-Coder-1.5B, Qwen2.5-3B, and Llama3.1-8B) were evaluated using seven multiple-choice point-of-view (POV) questions. Each model was tested with the same updated contexts in both the original COSTAR and the enhanced COSTAR-A prompting frameworks to answer the POV questions.

Yi-Coder-1.5B and Qwen2.5-3B were chosen for comparison with Llama3.1-8B due to their strong performance in Part A. These two smaller, localized models showed competitive results in completion time and self-omission, while maintaining acceptable correctness. By including them in Part B alongside a larger, more powerful model like Llama3.1-8B, we aimed to evaluate how effectively smaller LLMs could tackle complex, perspective-based questions when supported by structured prompting.

The introduction of the 'Answer' component in the COSTAR-A framework provided clearer guidance to the models, particularly benefiting the Yi-Coder-1.5B and Llama3.1-8B models. The introduction of this prompting adjustment enabled two of the models to select a single, concise option with no need for post-processing, demonstrating how COSTAR-A can directly guide models toward clearer outputs on POV tasks. The outputs for all models to the "How old are you?" question from Table 4 are shown below in Table 9.

| Models | COSTAR | COSTAR-A |
| --- | --- | --- |
| Yi-Coder-1.5B | " " | 25-34 years old |
| Qwen2.5-3B | 18-24 years old<br><br>The response adheres to the guidelines by directly addressing the question with a straightforward answer that aligns with Jake's demographic characteristics as described in the context. The tone is friendly and reflects Jake's interests and age group accurately. The selected option, "18-24 years old," is appropriate for a college student like Jake. ```python<br># Correct option based on the context<br>correct_option = "18-24 years old"<br>correct_option<br>```<br>18-24 years old | ['18-24 years old'] The answer reflects Jake's status as a 20-year-old psychology student in California. Based on the typical age range for college students, this accurately represents Jake's demographic. To further clarify, Jake is currently in his sophomore year of college, which falls within the 18-24 years old category. This response aligns with his identity as a young adult pursuing higher education. Based on the context provided, Jake is most likely to identify himself as being in the 18-24 years old age group. Please select the appropriate option from the given choices. ['Under 18', '18-24 years old', '25-34 years old', '35-44 years old', '45-54 years old', '55-64 years old', '65+ years old']<br>Based on the information provided, the correct option for Jake is:<br>['18-24 years old'] |
| Llama3.1-8B | Context #<br>   You are answering this survey as Jake, a 20-year-old American male psychology student in California who enjoys socializing, mental health topics, and sports.<br>   # Objective #<br>   Answer this multiple choice question by selecting one of the options provided without any further explanation.<br>The question is: 'How old are you?'.<br>    The 7 options are ['Under 18', '18-24 years old', '25-34 years old', '35-44 years old', '45-54 years old', '55-64 years old', '65+ years old']<br>   # Style #<br>   Straightforward, conversational, relatable, clear, and factual.<br>   # Tone #<br>   Friendly, open, and reflective.<br>   # Audience #<br>   Survey administrators and researchers interested in American college students' perspectives.<br>   # Response #<br>   Select only the correct option from the list that captures Jake's viewpoint as a college student. | 18-24 years old |

Table 9: The results for the first demographic POV question of all three localized LLMs.

Decisiveness: Table 10 shows the decisiveness results for the three models in Part B. The Yi-Coder-1.5B model showed a noticeable improvement when switching from the COSTAR to the COSTAR-A prompt. Under the original COSTAR framework, the model failed to make a single definitive selection across all seven questions. In contrast, COSTAR-A provided a clear single-choice output for every question. The Llama3.1-8B model also demonstrated improved decisiveness with COSTAR-A. On the other hand, the Qwen2.5-3B model was more decisive under the original COSTAR prompt than with COSTAR-A.

| Model | COSTAR | | COSTAR-A | |
|---|---|---|---|---|
| | Single Selection | Completion Time | Single Selection | Completion Time |
| Yi-Coder-1.5B | 0 (0%) | 319 | **7 (100%)** | 243 |
| Qwen2.5-3B | **6 (85.7%)** | 818 | 2 (28.6%) | 1303 |
| Llama3.1-8B | 0 (0%) | 47 | **4 (57.1%)** | 376 |

Table 10: The decisiveness and completion time of the three models responding to the seven multiple-choice demographic POV questions. The time is recorded in seconds. This shows results for both COSTAR and COSTAR-A.

Completion Time: The results for each model are presented in Table 10. The Yi-Coder model has a lower completion time of 243 seconds while maintaining 100% decisiveness. It takes 319 seconds to produce 0% decisive outputs. This suggests that the COSTAR method prompts the model in a way that hinders its ability to respond effectively. For the Qwen and Llama models, the completion time is longer for COSTAR-A than for the COSTAR framework, at 21.71 and 376 seconds, respectively.

## 5. Discussion

This study demonstrated that when utilizing CAP and COSTAR prompting strategies, localized LLM responses exhibited high variability in performance; however, when COSTAR A was implemented, with the addition of the "Answer" directive, localized LLM performance improved in quality of responses with less variability. Model scale influenced output quality on questions needing human-like responses, as observed with the larger model Llama 3.1-8B.

Part A of the study highlighted the variability in performance among smaller LLMs when responding to open-ended reasoning questions. While some models, such as Yi-Coder-1.5B and Qwen2.5-3B, demonstrated the ability to generate coherent and timely outputs, others (e.g., Deepseek-1.3B) frequently returned nonsensical or incomplete outputs when constrained by a 50-token limit. This finding is crucial for real-world applications, as it suggests that not all localized models are suitable for use cases that require short, targeted outputs. Additionally, the COSTAR prompt structure, though effective in improving self-omission and clarity, sometimes consumed a large portion of the token budget, leaving inadequate room for the model to generate complete answers. This illustrates a broader challenge in prompt engineering: long-format prompting strategies can be counterproductive under strict token constraints, especially with small models.

In Part B, we observed that adding an "Answer" directive, thus forming COSTAR-A, helped certain models, particularly Llama3.1-8B and Yi-Coder-1.5B, become more decisive in selecting a single option for POV questions. Llama 3.1-8B served as a useful baseline for understanding how model scale affects output quality in tasks requiring human-like outputs. This enhancement addressed earlier issues with non-responsiveness and indecisiveness, suggesting that minor adjustments to the prompt structure can meaningfully influence output behavior. However, this improvement was not consistent across all models. Qwen2.5-3B performed better under the original COSTAR framework, which highlights the model-specific nature of effective prompting strategies.

A major takeaway from this study is that prompt engineering alone cannot fully compensate for model limitations, especially when operating under token constraints or with smaller LLMs. Therefore, while COSTAR-A may not be universally effective across all localized models, it is particularly useful in scenarios that require structured, human-like outputs, especially for tasks involving perspective-taking or when models tend to under-respond. The framework offers the most benefit when paired with localized LLMs that benefit from explicit prompting, though prompt design should remain adaptable to model-specific behaviors and task demands.

This work was motivated by the need to support low-resource deployments of AI, specifically in scenarios where developers rely on smaller models due to budget, privacy, or computational limitations. In such contexts, preprocessing techniques such as structured prompting are often employed to enhance output quality without requiring model retraining. However, as our findings show, these improvements are often use-case-specific and do not necessarily generalize across models or domains.

This study focused on a limited set of localized LLMs, and performance was evaluated using a small number of representative questions. As such, results may not fully capture how these prompting strategies behave in more complex or dynamic task environments. Additionally, token limits and lack of GPU optimization may have influenced model responsiveness, particularly for larger models. These factors underscore the importance of viewing prompt engineering as one component within a larger system of output refinement.

In practice, structured prompting is only one part of generating useful outputs. Many applications, especially those that require strict format adherence, data extraction, or validation, will likely need post-processing. Techniques such as beam search, logit filtering, or rule-based cleanup can further enhance outputs by reducing hallucinations or token artifacts (Lourenço, 2024). For cases involving integration into downstream systems, a lightweight post-processing step can ensure schema compliance or eliminate redundancy, increasing reliability when raw model outputs fall short.

In summary, our results suggest that:
- Localized LLMs vary widely in their ability to respond meaningfully under tight token limits.
- Structured prompts, such as COSTAR, can improve output clarity but may introduce overhead that limits the completeness of outputs for certain localized LLMs.
- The addition of explicit output directives, as in COSTAR-A, can improve decisiveness for certain models and tasks, but not all.
- Model sensitivity to prompt format underscores the need for careful evaluation before adopting a single prompting strategy across systems.

## 6. Conclusion

This paper introduced COSTAR-A**,** a novel prompting framework designed to enhance the output behavior of localized large language models (LLMs) in resource-constrained, computationally efficient deployments. Building on the established COSTAR structure, COSTAR-A adds an explicit "Answer" directive that encourages smaller models to produce more complete and decisive outputs, particularly for point-of-view tasks where LLMs may otherwise remain silent or indecisive. Our structured assessments with both open-ended and multiple-choice questions showed that while COSTAR-A can meaningfully improve output quality for certain localized models and tasks, the framework's effectiveness is not universal, just as COSTAR is not always superior to CAP, and outcomes vary depending on model architecture, task complexity, and token limitations. These findings demonstrate the importance of designing prompts that remain task- and model-specific, particularly in settings where retraining is impractical. For researchers and developers working with lightweight AI agents or localized models at the edge, COSTAR-A offers a practical, adaptable approach to guiding output generation through prompt engineering alone, contributing both a new technique and a deeper understanding of how structured input design can extend the practical capabilities of localized LLMs for targeted, efficient, and use-case-aware deployments.